\documentclass[sigconf]{acmart}
\renewcommand\footnotetextcopyrightpermission[1]{}
\pdfoutput=1 

\usepackage{times}
\usepackage{latexsym}
\usepackage{graphicx}
\usepackage{tabularx}
\usepackage{tablists}
\usepackage{microtype}
\usepackage{multirow}
\usepackage{amsmath}

\usepackage{amssymb}
\usepackage{pdfpages}
\usepackage{booktabs} 
\usepackage{fancyhdr}
\AtBeginDocument{%
  \providecommand\BibTeX{{%
    \normalfont B\kern-0.5em{\scshape i\kern-0.25em b}\kern-0.8em\TeX}}}

\begin{document}

\fancyhead{}



\title{Attract me to Buy: \\
Advertisement Copywriting Generation with Multimodal Multi-structured Information}

\author{Zhipeng Zhang$^{1,2}$ \quad Xinglin Hou$^{2}$ \quad Kai Niu$^{1}$  \quad Zhongzhen Huang$^{1}$ \quad Tiezheng Ge$^{2}$ \\
\quad Yuning Jiang$^2$  \quad Qi Wu$^{3}$\quad Peng Wang$^{1,*}$} 

\affiliation{
 \institution{\textsuperscript{\rm 1}School of Computer Science, Northwestern Polytechnical University} 
 \institution{\textsuperscript{\rm 2}Alibaba Group}
 \institution{\textsuperscript{\rm 3}University of Adelaide, Australia }
 \country{}
 }
 
\email{{zhipengzhang, actor}@mail.nwpu.edu.cn, {peng.wang, kai.niu}@nwpu.edu.cn}
\email{{xinglin.hxl, tiezheng.gtz, mengzhu.jyn}@alibaba-inc.com}
\email{qi.wu01@adelaide.edu.au}

\begin{abstract}
Recently, online shopping has gradually become a common way of shopping for people all over the world. Wonderful merchandise advertisements often attract more people to buy. These advertisements properly integrate multimodal multi-structured information of commodities, such as visual spatial information and fine-grained structure information. However, traditional multimodal text generation focuses on the conventional description of what existed and happened, which does not match the requirement of advertisement copywriting in the real world. Because advertisement copywriting has a vivid language style and higher requirements of faithfulness.  Unfortunately, there is a lack of reusable evaluation frameworks and a scarcity of datasets. Therefore, we present a dataset, E-MMAD (e-commercial multimodal multi-structured advertisement copywriting), which requires, and supports much more detailed information in text generation. Noticeably, it is one of the largest video captioning datasets in this field.  Accordingly, we propose a baseline method and faithfulness evaluation metric on the strength of structured information reasoning to solve the demand in reality on this dataset. It surpasses the previous methods by a large margin on all metrics. The dataset and method are coming soon on \url{https://e-mmad.github.io/e-mmad.net/index.html}.
\end{abstract}

\maketitle
\section{Introduction}
\label{introduction}

Nowadays, online shopping has become one of the main ways for people to shop, such as Taobao, Amazon. The product advertisement is often an important factor in people's shopping. Wonderful  product advertising can attract people's attention and promote sales.  Commodity advertisement copywriting\cite{duan2021query} presents commodities in a more concise and intuitive way, which is convenient for people to search and shop. Different from conventional text generation \cite{chen2019few,lu2018neural}, commodity advertisement copywriting often has vivid language style and flexible grammar. Meanwhile, it also needs to comprehensively consider multimodal information and fine-grained multi-structured information\cite{song2021structural} parameters of commodities, which results in sellers often need to spend a lot of manpower, time and money on elaborate design to produce high-quality advertising copywriting. We named this more challenging problem  as multi-modal e-commerce advertisement copywriting generation.

\begin{figure}[tb]

\includegraphics[width = 1\linewidth]{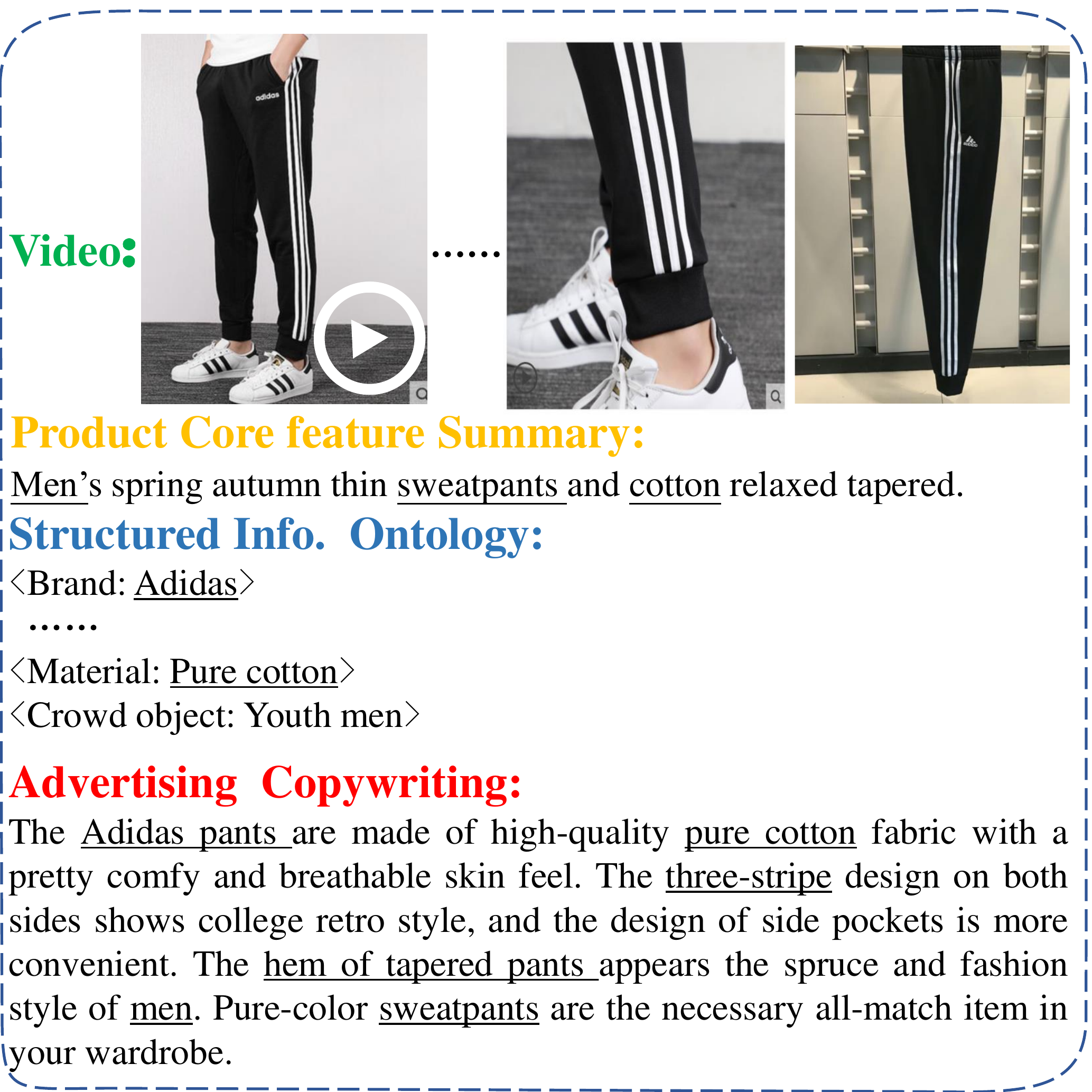}

\caption{An illustration of our dataset. The four different parts of our dataset, from top to bottom are product information (commodity displaying video, the product core features summary, structured information) and commodity advertising description. We use multimodal multi-structured information to assist in generating a semantically richer copywriting. The underlined words are closely related to the advertising copywriting, and are also important in terms of faithfulness.}
\label{fig:intro}
\end{figure}

Multimodal product advertising text generation is different from video Caption \cite{xu2016msr, wang2019vatex}, natural language generation \cite{chen2019few,chi2020cross}, etc. The information sources needed for the commodity copywriting  are diversified, which are closely related to the video display of commodity, commodity structured Info. Ontology and commodity core feature summary.  For example, as shown in Figure~\ref{fig:intro}, the product video shows the spatial and visual color impact information of the product, while the structured information of the attribute word lists shows the precise fine-grained information of the product, such as brand . The product core features summary (unstructured information), summarizes the core features of the product in a general way. Through multimodal fusion, structured and unstructured information is combined to directly generate high-quality fine-grained text. This is also consistent with the fact that in real life, text generation of commodity advertisements is closely related to multi-modal information sources, and details of different modes are considered comprehensively. Therefore, how to extract the required information from multimodal and fully integrate is one of the challenges in this task.

The other of the challenges is faithfulness\cite{2022Faithfulness}. For commodity advertising copywriting, its fine-grained key information should greatly reduce factual errors of principle. The existing Video caption methods may achieve good results under the traditional regularized language paradigm of video caption, but often make factual errors in the grammatically flexible advertisement copywriting. For example, for an Adidas shoe, a model-dependent output branding error cannot occur because a similar Nike shoe appears in the training set. Faithfulness is an extremely critical core issue.  This information from product structural attributes can make the description precise and reliable.  Meanwhile, the current metrics\cite{zhang2019bertscore,2019Towards,papineni2002bleu,lin2004rouge,vedantam2015cider} cannot accurately calculate the accuracy of attribute words in the real-world advertisement application, for instance, BERTScore\cite{zhang2019bertscore} is based on the knowledge pre-train and KOBE\cite{2019Towards} is also based on N-gram\cite{papineni2002bleu} for lexical diversity. To this end, we proposed a corresponding metric that makes full use of structured information, hard homologous metric, for auditing the faithfulness of measured structured information.

To address these challenges, we elaborately collect a large-scale e-commercial multimodal multi-structured advertising dataset for multimodal text generation research. To support in-depth research, we collect a rich set of product annotations. Our dataset consists of 120,984 product instances in both Chinese and English, in which each instance has a product video, a product core feature summary, structured information ontology and advertising copywriting.  In response to the realistic demand for advertising generation, we propose the multimodal information fusion module and generation decoder module which make full use of the rich information. In faithfulness\cite{2022Faithfulness}, we propose Conceptualization Operations to conceptualize complex and diverse information in real life as ontology, which can help model to face various words. An ontology models generalized data, that is, we take into consideration general objects that have common properties and not specified individuals.   The proposed network leads to a significant improvement over existing practical application methods, on our constructed dataset. Dataset and code will be available at our Website.

In summary, our contributions concentrate on the following aspects:\\
(1) We introduce a fresh task: e-commercial multimodal advertisement copywriting generation.   A new large-scale high-quality and reliable e-commercial multimodal advertising dataset is introduced, which  requires and supports multimodal fusion and faithfulness accuracy. It is also one of the largest video-text datasets in this field. E-MMAD is collected from human real life scenes and carefully selected so that it is qualified to meet the needs of this task.\\
(2) We propose a simple but effective strong baseline method to solve the challenges  in reality. Our approach achieves the Top-1 accuracy in faithfulness and other  metrics, outperforming existing reusable methods.\\
(3) As for the faithfulness of advertising copywriting, we propose the hard  homologous metric. This metric can correspond to the advertisement copywriting source.

\section{Related Work}
\label{related_work}
\subsection{Multimodal video-text generation datasets}
There are various datasets for multimodal video-text generation that cover a wide range of domains, such as movies \cite{rohrbach2015dataset,rohrbach2017movie}, cooking \cite{das2013thousand,zhou2018towards}, and activities \cite{xu2016msr,carreira2018short}. MSR-VTT \cite{xu2016msr} is a widely-used dataset for video captioning, which has 10,000 videos from 257 activities and was collected in 2016. MSVD \cite{chen2011collecting} was collected in 2011, containing 1970 videos. ActivityNet \cite{caba2015activitynet} has 20,000 videos but is used for Dense Video Captioning \cite{krishna2017dense,iashin2020multi}, which means to describe multiple events in a video. TVR \cite{lei2020tvr} is collected from movie clips whose text is mainly character dialogue. Vatex \cite{wang2019vatex} is a famous dataset released in 2019, whose caption is written by batch manpower. Poet\cite{zhang2020poet} is an e-commerce video feature dataset containing two small raw datasets BFVD and FFVD. The data was downloaded directly from the Internet and not carefully filtered by human multimodal alignment. As shown in Table~\ref{comparison}, we generated a larger video dataset after a lot of time and manpower screening. In addition, the struct info ontology that we emphasize is carefully selected and generated by us to solve the real faithfulness problem\cite{2022Faithfulness}, not from the rough data of the network. This part will be used to solve the faithfulness in the advertisement. We find that the advertising caption includes a lot of structured information in fact.  Compared with some mainstream datasets in Table~\ref{comparison}, our dataset also provide an additional product structured information.

\subsection{Video Captioning Approaches}
Video caption/description is one of the important tasks in multimodal text generation\cite{iashin2020multi}. Early video caption methods are all based on templates \cite{mitchell2012midge,krishnamoorthy2013generating,tang2002spatial}. However, sentences made in this way tend to be rigid. The sequence-to-sequence model \cite{venugopalan2015sequence} is a classic work, which includes an encoding phase and a decoding phase. After CNN extracts the image features of the video frames, an image feature is sent to the LSTM for encoding at each time step and text will be generated in the decoding stage\cite{jia2015guiding}. Some of the popular practices recently are based on data-driven  \cite{zhang2021open} and transformer-based mechanisms  \cite{yang2019non,zhou2018end,lei2020mart}. MART  \cite{lei2020mart} can produce more coherent, non-repetitive, and relevant text to enhance the transformer architecture by using memory storage units\cite{pei2019memory,xu2015show}. Vx2text \cite{lin2021vx2text} uses multimodal inputs for text generation. They use a backbone \cite{tran2018closer,ghadiyaram2019large} model to transform different modalities information to natural language and then the problem turns to natural language generation. Recently, there are works\cite{hu2019hierarchical, yang2017catching,wang2018spotting} extracting object-level features in representing the videos for video caption.  Although good progress has been made by them, the original information of the modal is not fully utilized and integrated. Moreover, they may achieve good results under the traditional regularized language paradigm of video caption, but often make factual errors in the grammatically flexible advertisement copywriting. 

\vspace{-0.1cm}
\section{Datasets}
\label{data}

In this section, we will introduce our dataset in detail, including the statistic analysis, collecting process, and comparison.

\subsection{Data Collection}

\textbf{1) Dataset sources.} Our dataset sources are the Chinese largest e-commerce website shopping platform\footnote{\url{www.taobao.com}}, from which we have collected nearly 1.3 million commodity examples with structured information. It comprised more than 4,000 merchandise categories to guarantee the diversity of the dataset, such as clothes, furniture, office supplies, etc. The information of each commodity data sample includes structured information, merchandise displaying video, product core feature summary  and commodity advertising description. Different from previous works \cite{wang2019vatex,xu2016msr,chen2011collecting}, the sources of datasets are derived from what merchants themselves numerously design and select, which comply with the standard rules of the authenticity of product advertisements and are supervised by false product advertising rules of \textit{Taobao}. This can avoid the templated sentences and vedios.
Specifically, videos visually display the commodity performance and application. So the E-MMAD are limited to e-commerce aspect.
In addition, we fully consider ethical privacy issues to ensure that the dataset has no potential negative effects and legal issues  \cite{gebru2018datasheets}. All data is collected in \textit{Taobao} shopping platform, which is a public platform for the general public. All information, even the characters in the video, is ensured to comply with Taobao laws\footnote{\url{https://rule.taobao.com}} including personal privacy, legal prohibitions, false information, protection of minors and women, and so on.

\begin{figure}[htb]
\setlength{\belowcaptionskip}{-0.3cm}
\scalebox{0.5}{
\includegraphics[width=\textwidth]{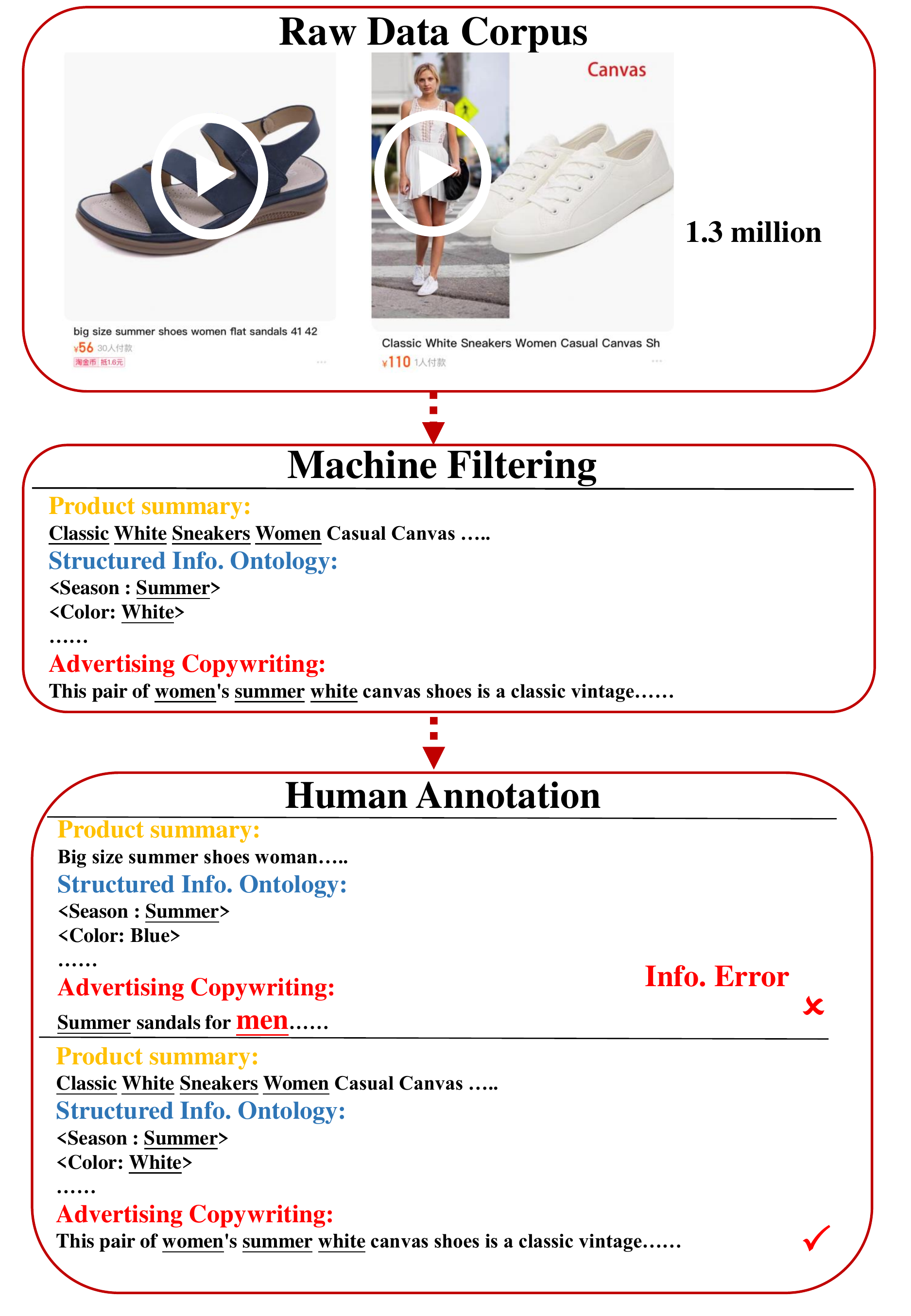}
}
\caption{The process of creating a dataset, including machine filtering, manual post-filtering, etc. and data specification of the dataset. Each set of data needs to be carefully filtered and annotated manually in order to produce high-quality multimodal dataset.}
\label{fig:regular}

\end{figure}

In consideration of data and ethics, we perform programmatic screening and manual cleaning again in accordance with the established data cleaning rules. Figure~\ref{fig:regular} shows our data collection process.

\textbf{2) Data filtering.} The intention for data filtering is to determine whether the product advertising description is closely related to the product displaying video, and whether the structured information of the product is in accordance with the composition of the product advertising description and ethical considerations. The product attributes structured information and product displaying video will be valid only if human being can write similar product advertising descriptions with them. We use programs to screen and judge at first. 
Our screening basis is the proportion of structured information words in the product advertising description. When the proportion is up to \textit{n} words or more, the data will be reserved as valid data. After copywriters' continuous attempt to generate advertising descriptions with structured information words that account for different proportions, we finally determine the structured information with more than five words in the product advertising description as valid data.\\
By virtue of this, we respectively test different groups of random data to formulate screening and judgment rules. Multiple copywriters tested and discussed to make the manual evaluation criterion several times.  Finally, different testers sample 100 examples randomly according to the judgment rules of supplementary, and the pass rate is mostly about $60\%$. In this case, we validate the manual screening rules and draw the conclusion that random subjective factors hardly have any influence. So far, the manual data screening and judging rules have been formed, as is shown in Dataset Supplementary.

\textbf{3) Data annotation.}   We invited 25 professional advertising copywriters as data screening and annotation staff to conduct manual screening under the rules in Dataset Supplementary and The Toronto Declaration\footnote{\url{https://www.torontodeclaration.org}} . Manual screening of all data also ensures that each piece of data complies with the Toronto Declaration and \textit{Taobao} laws to protect gender equality, racial equality, etc. In order to ensure the reliability of the data, we use the following two methods to sample and verify: 
(1). Add verification steps. We will send back samples that have been annotated right answers to annotators from time to time to check their work quality.
(2). Multiple people Choices. The data is sent to different people randomly. Only if the answers of all people are consistently passable, can this data be qualified. Finally, 120,984 valid data has been generated.

Simultaneously, we also translate the filtered valid data into English so that both Chinese and English versions can be provided in the dataset. To ensure the quality of the English version, we use the WMT 2019 Chinese-English translation champion, baidu machine translation. We also monitor the translation quality in the manual screening section, such as random checking in batch translation and back translation comparison. 

After people's diligent work of manual data labeling and cleaning, there are 120,984 eligible data selected finally.

\begin{figure}[tb]
\setlength{\belowcaptionskip}{-0.5cm} 
\centering
\scalebox{0.48}{
\includegraphics[width=\textwidth]{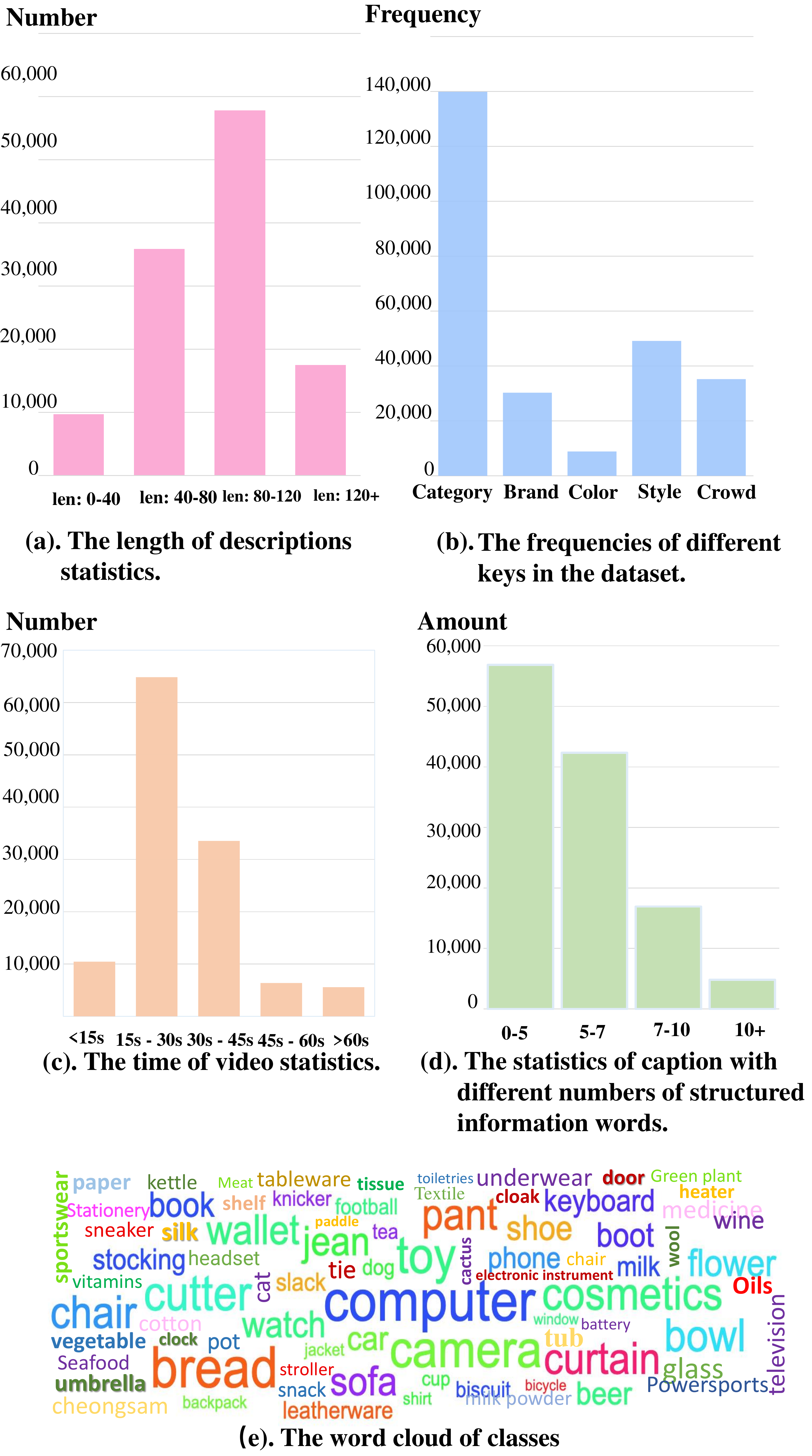}
}
\caption{Statistics about the five different forms of data in our dataset. The data statistics are presented in terms of video, structured information, caption, and the main classes of the dataset contained, respectively.}
\label{fig:statisc}
\end{figure}

\subsection{Dataset Analysis}
\begin{table*}[htb]
\caption{Comparison with other datasets. \emph{\# Videos}, \emph{Average Time}, \emph{Caption Length}, \emph{\# Classes} respectively represent the total number of videos in the dataset, the average video time in the dataset, the average length of the captions in the dataset and the number of instance types in the dataset. \emph{Input Modality} indicates the input of the dataset, e.g. from Video to Text, Multimodal to Text. \emph{Structure info.} means whether the dataset contains structured information. There are 3,876 keys of the structured information in E-MMAD dataset. en means English version dataset and zh means Chinese version dataset.}
\centering 
\begin{tabular}{l|c|c|c|c|c}
\toprule
\textbf{Datasets} & \textbf{\#Videos} &\textbf{\begin{tabular}[c]{@{}c@{}}Average\\ Time\end{tabular}}&  \textbf{\begin{tabular}[c]{@{}c@{}}Caption\\ Length\end{tabular}} & \textbf{\#Classes} & \textbf{Input Modality}\\ 
\midrule
MSR-VTT (en) \cite{xu2016msr} & 10,000  & 14.8s& 9 &257 &Video    \\
MSVD (en)\cite{chen2011collecting}     & 1,970& 9.0s &  8 & - &Video     \\
TVR (en) \cite{lei2020tvr} &21,800 &9.0s &13&-& Video-query    \\

VaTEX (en/zh) \cite{wang2019vatex}    & 41,269 &10.0s &  15/13& 600&Video     \\
FFVD (zh) \cite{zhang2020poet}    & 32,763 &27.7s &  62& - &Video - Attribute  \\
BFVD (zh) \cite{zhang2020poet}    & 43,166 &11.7s &  93& - &Video - Attribute   \\
E-MMAD (en/zh) &\textbf{120,984} & \textbf{30.4s} & \textbf{97/67}&\textbf{4,863}&\textbf{Video - Summary - Structure info. } \\
\bottomrule
\end{tabular}

\label{comparison}
\end{table*}

We make an elaborate analysis on these valid data and the result is shown in Figure~\ref{fig:statisc}. In addition to this, Figure~\ref{fig:statisc} reveals the distribution of the product videos' duration and advertising descriptions. 

By Table~\ref{comparison} comparison, we can find that our product advertising descriptions are not only at least twice longer than others, but also root in more vivid and realistic ones used in practice. The whole statistics about the structured information in our dataset is displayed in Figure~\ref{fig:statisc} (d).  What's more, there exist average 21 structured information words in each sample and 6.2 words of them are finally displayed in its product advertising copywriting. The \textbf{(e)} shows the abundance of our datasets source classes.

\subsection{Dataset Comparison}
In Table~\ref{comparison}, we make a comparison between our dataset and others from the following several perspectives: dataset scale, dataset diversity and dataset reliability.

\textbf{(1). Dataset scale:} As shown in Table~ \ref{comparison}, the size of our E-MMAD is the largest multimodal dataset among those we have already known so far, with the longest video duration and text length, and the richest structured information in the dataset.

\textbf{(2). Dataset Diversity:} In terms of types, our dataset consists of 4,863 categories. Our dataset is also available in Chinese and English two versions, to support multi-language research, which cannot be satisfied by a single language dataset. At the same time, our Chinese and English corpus is richer in vocabulary, which is from the real-world more natural and diversified language style.

\textbf{(3). Dataset Reliability:} Compared with other manual batch-written descriptions\cite{wang2019vatex} and mechanically generated data, our data annotation is derived from the real society. Each of them is an exclusive description genuinely written by corresponding store. Besides, the videos in our dataset are from the real product shooting scene, other than clips from movies or others. We firmly believe that only resorting to reliable dataset, can we train models better. Therefore, we invest considerable amount of manpower and time in order to promote our dataset quality.

\begin{figure*}[h]

\centering
\scalebox{1}{
\includegraphics[width=\linewidth]{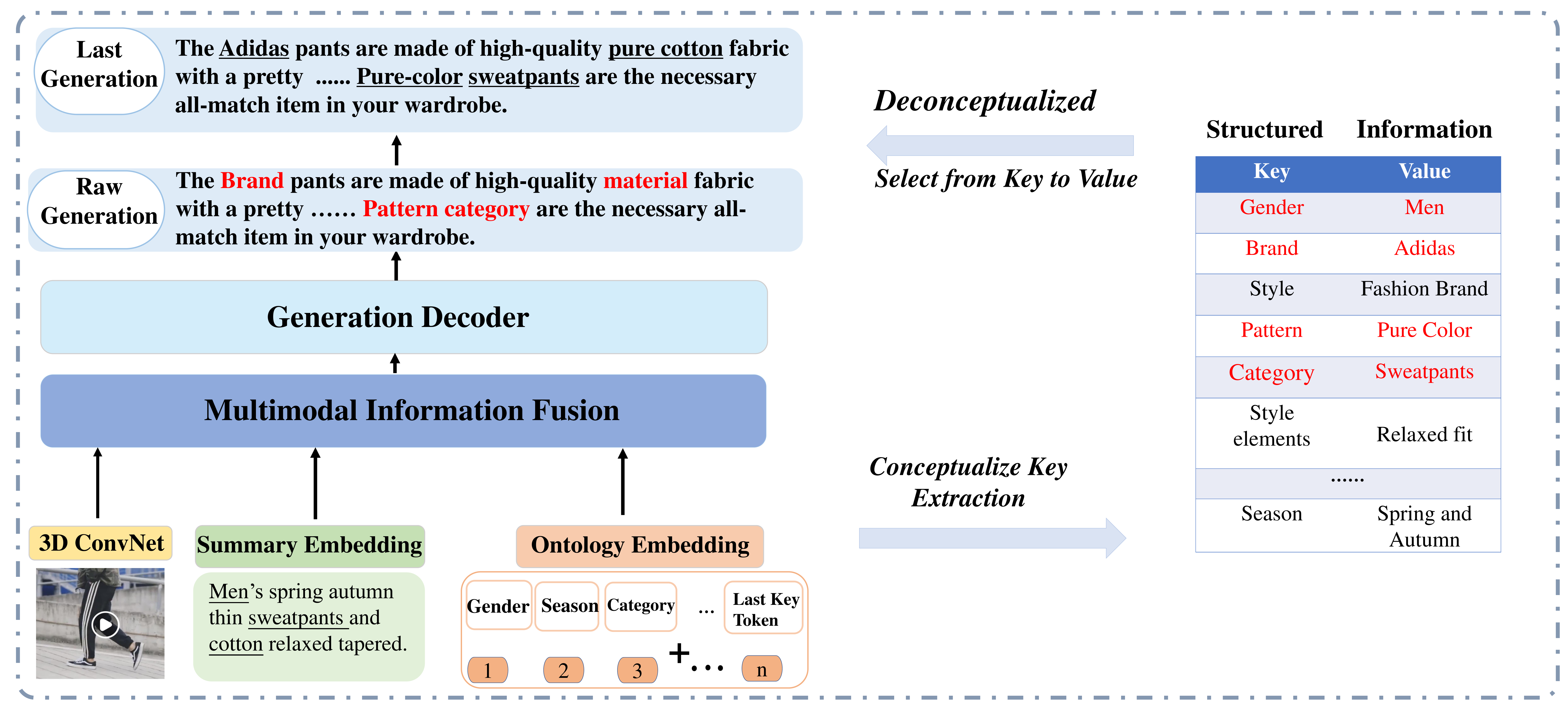}
    }
\caption{The overall architecture of our model, which contains three main parts: the representation for multimodal information, the multimodal fusion module based on self-attention and the generation decoder module on the basis of \cite{radford2019language}. According to the Key-Value, the used Structure information words are conceptualized as ontology to face the various words such as assorted brands in real life.\\ }
\label{fig:framework}

\end{figure*}

\subsection{Dataset Significance}
To the extent of our knowledge, the dataset we propose is one of the largest multi-modal dataset so far, and the information involved is also the most diverse, which can better optimize and improve the performance of multi-modality models and promote their generalization ability to adapt to different scenarios in real world. For subsequent work, with the abundant and diverse information involved, our dataset can be dedicated to several multi-modality domain tasks, such as Video Retrieval \cite{lei2020tvr,dzabraev2021mdmmt}, Product Search \cite{chang2021extreme} and so on.

\section{Method}
\label{method}
In this work, we present a novel approach called the Multi-modal Fusion and Generation algorithm as shown in Figure~\ref{fig:framework}, which extracts feature representations from three sources: the product core feature summary, structured information and the displaying video's frames and fuse them to generate captions. Faced with various information words, our model uses ontology, a method of conceptualizing information. That is to pre-process the various data, conceptualize and extract information from the complex information words to Key as highly conceptual network features.  For the restoration of complex information in the generation phase, we only need to perform the inverse conceptualization operation at the end.
\subsection{Conceptualization}\label{Conceptualization}
During the training process, we pre-conceptualize the true product descriptions. The formula is as follows:
\begin{gather}
\setlength{\belowdisplayskip}{3pt}
Values_{gr} = SI.values \bigcap GR.tokens; \\
k_{gr} \in SI.keys;\\
\begin{split}
token_{gr}&\rightarrow k_{gr}, \\
\forall \  token_{gr} &\in Values_{gr}. \\
\label{con:concept}
\end{split}
\end{gather}

In the generation process, the raw copywriting with conceptualized information generated by the model is de-conceptualized to obtain the final copywriting. The de-conceptualization is as follows:
\begin{gather}
Values_{rc} = SI.keys \bigcap RC.tokens;\\
v_{gr} \in SI.values;\\
\begin{split}
rc\_token &\rightarrow v_{gr}, \\ \forall \  rc\_token &\in Values_{rc}. \\ 
\label{con:deconcept}
\end{split}
\end{gather}
In equation~(\ref{con:concept}) (\ref{con:deconcept}), $A\rightarrow B$ means replacing token $A$ with token $B$. $A\in C$ means token $A$ is an element of set $C$. $GR.tokens$ and $RC.tokens$ are the sets of corresponding n-gram phrases in ground truth and raw caption, respectively. $SI.values$ and $SI.keys$ respectively correspond to the sets of keys and values in the structured information. In terms of the model input, the ontology of the structured information part is conceptual value words. An ontology models generalized data, that is, we take into consideration general objects that have common properties and not specified individuals. By this, the 3,876 types of Keys represent the various information words as the highly conceptual feature input. We also reference the summary as the basis to determine the priority position of each key according to the order in which the structured information appears in the summary.
\subsection{Representation}
\textbf{Textual Information.} Given a product core feature summary as a list of \textit{K} words, conceptualized product structure information as a list of \textit{N} keys, we embed these words and keys into the corresponding sequence of \textit{d}-dimensional feature vectors using trainable embeddings \cite{zhang2021cpm,devlin2018bert}. In addition, since the keys of structured information are prioritized, we use \emph{position embedding} to represent the priority position of the keys. \\
\textbf{Visual Information.} Given a sequence of video frames/clips of length \textit{S}, we feed it into pre-trained 3D ConvNet\cite{ji20123d} to obtain visual features $V=\left\{v_{1}, v_{2}, \ldots, v_{K}\right\} \in \mathbb{R}^{S \times d_{v}}$ , which are further encoded to compact representations $R \in \mathbb{R}^{S \times d}$. The \textit{Visual Embedding Layer} can be formalized as following:
\begin{gather}
    f_{VEL}(v) =\operatorname{BN}(g \circ \bar{v}+(1-g) \circ \hat{v}); \\
    \bar{v} =W_{1} v^{\top}; \\
    \hat{v} =\tanh \left(W_{2} \bar{v}\right);\\
    g =\sigma\left(W_{3} \bar{v}\right).
\end{gather}
where $BN$ denotes batch normalization, $\circ$ is the element-wise product, $\sigma$ means sigmoid function, $W_{1} \in \mathbb{R}^{d \times d_{v}}$ and $\left\{W_{2}, W_{3}\right\} \in \mathbb{R}^{d \times d}$ are learnable weights.

\subsection{Multimodal Fusion}
After embedding all information from each modality as vectors in the \emph{d}-dimensional joint embedding space, we use a stack of \emph{L} transformer layers with a hidden dimension of \emph{d} to fuse the multi-modal information consisting of a list of all $K+N+S$ modalities from $\left\{v_{S}^{\text {frames }}\right\}$ $\left\{v_{K}^{\text {words}}\right\}$ and $\left\{v_{N}^{\text{keys}}\right\}$. Through the self-attention mechanism in transformer, we can model inter- and intra- modality context. The output from our Multimodal Information Fusion module is a list of \emph{d}-dimensional feature vectors for entities in each modality, which can be seen as their interrelated embedding in multimodal context. In this work, the parameters chosen for our the module are consistent with the parameters of \cite{devlin2018bert} (L=12, H=768, A=12), where \emph{L}, \emph{H}, \emph{A} represents the number of layers, the hidden size, and the number of self-attention heads respectively.

\begin{table*}[htb]
\caption{Performance ($\%$) comparison with our proposed model and others. The NACF + multi-input means that we concat the structured information and summary with video feature directly as input. On the premise of fair comparison, the following methods are relatively classic and available, which are applicable on E-MMAD by our objective attempts. }
  \label{reslut_compare}
  \centering

\begin{tabular}{c|c|l|c|c|c|c|c|c}
\hline
\multicolumn{1}{l|}{Version} & Input                       & Method    & Bleu1         & Bleu2         & Bleu3        & Bleu4        & Rouge\_L      & CIDEr         \\ \hline
\multirow{5}{*}{en}          & Text                        & NLG \cite{chen2019few}       & 13.6          & 6.8           & 3.1          & 1.9          & 13.0          & 10.1          \\ \cline{2-2}
                             & Video                       & NACF \cite{yang2019non}      & 18.9          & 7.9           & 3.9          & 2.2          & 15.3          & 14.8          \\ \cline{2-9} 
                             & \multirow{3}{*}{Multimodal} & NACF + multi-inputs    & 20.0          & 8.5           & 4.3          & 2.4          & 17.8          & 18.5          \\
                             &                             & TVC \cite{lei2020tvr}      & 21.3          & 12.4          & 6.2          & 3.7          & 19.3          & 22.5          \\
                             &                             & \textbf{Ours (en)} & \textbf{25.0} & \textbf{16.6} & \textbf{9.6} & \textbf{7.2} & \textbf{25.3} & \textbf{29.1} \\ \hline
\multirow{2}{*}{zh-CN}       & Text                        & CPM (zh) \cite{zhang2021cpm}  & 7.9           & 4.6           & 1.1          & 0.5          & 7.2           & 8.3           \\ \cline{2-2}
                             & Multimodal                  & \textbf{Ours (zh)} & \textbf{11.6} & \textbf{6.5}  & \textbf{4.4} & \textbf{2.2} & \textbf{12.5} & \textbf{15.3} \\ \hline
\end{tabular}

\end{table*}

\subsection{Generation Decoder}
Our model's decoder is a  Transformer decoder, which follows the mainstream model architecture of \cite{chen2019few,radford2018improving}. The decoder accesses multimodal fusion outputs at each layer with a multi-head attention \cite{vaswani2017attention}. Specifically, the decoder applies a multi-headed self-attention over the caption textual feature. After that, the position-wise feed forward layer was used to produce a distribution probability of each generation tokens for the final generated caption. There is a description of part of the formula for the decoder module: 
\begin{gather} 
    h_{0} =V^{\text{cap}}\cdot W_{t}+PE \cdot W_{p}; \\
    h_{l} =\text { Trans\_Block }\left(h_{l-1}\right);  \\
    P(w) =\operatorname{Softmax}\left(h_{n} W_{e}^{T}\right);\\
    PE_{(pos, 2i)} =\sin \left(pos / 10000^{2 i / d_{\text {model }}}\right); \\
    PE_{(pos, 2i+1)} =\cos \left(pos / 10000^{2 i / d_{\text {model}}}\right);
\end{gather}
where $V^{cap}=\left\{v_{1}, v_{2}, \ldots, v_{x}\right\}$ is the textual vector of caption, \emph{n} is the number of layers, $ \forall l \in[1, n]$, and $W_t$, $W_{p}$ is the learnabale weight for caption embedding feature and position encoding respectively. $Trans\_Block$ represents a block of the decoder in the Transformer \cite{vaswani2017attention}. We refer to \cite{radford2019language} as the model decoder architecture.


\section{Experiments}
\label{Experients}
In this section, we will show a series of experiments of our proposed model, including ablation studies, comparison experiments and metric.
\subsection{Implementation Details}
All the experiments are conducted on Nvidia TitanX GPUs. The proposed model is implemented with PyTorch.  For the representations of videos, we follow \cite{yang2019non} for fairness and opt for the same type, first extract 3D features with 2048 dimensions\cite{hara3dresnets}. For generation decoder, we use \emph{<sep>} to separate the input from the ground truth of caption. We adopt diverse automatic evaluation metrics to compare with other model: BLEU \cite{papineni2002bleu}, Rouge-L \cite{lin2004rouge}, and CIDEr \cite{vedantam2015cider}.  Since the major information captured by each model is different, the key information component of the generated caption will not be the same, but it is cognitive at the semantic level, so the CIDEr evaluation metric will have a relatively large fluctuation. Our model makes full use of structured information so that the generated caption can include most of the major detail information. However, the faithfulness degree is not well reflected in Table~\ref{reslut_compare}.   

\subsection{Comparison with Other Approaches}
\label{compare_sota}

During the comparison experiments, we uniformly divided the Chinese (zh-CN) and English (en) versions of our dataset into training set, validation set and test set in the ratio of 6:2:2 for training and testing. Since the current mainstream models do not use multimodal data for captioning, we use unimodal data for captioning on some classic and available methods, such as video caption, nlg, etc. For the sake of fairness of comparison, we simply modify the input part of the above experimental model to accommodate multimodal data. As we can see from Table~\ref{reslut_compare}, the comparison of the results before and after the model modification shows that multimodal data can substantially improve text generation tasks. It indicates that multimodal information indeed helps captioning by modal information between the mutual enhancement. As shown in Table~\ref{reslut_compare} our algorithm achieves a better performance than other methods because our model makes better use of multimodal data in the means of fusing different modalities and multi-structured information to reason.

\subsection{Ablation studies}
\textbf{Multimodal Input.} We perform ablation studies based on changing the input components of our proposed model as a way to validate the importance of our proposed dataset containing multi-structured information. As shown in Table~\ref{input_result}, we analyze the gap between the generated caption of the model and the real merchandise advertising description in the absence of partial information. As we can see, the absence of any of the three input components significantly degrades the final generated caption result. From our analysis of the generated caption samples, we can conclude that: 1) the lack of structured information will make the generated caption less informative, rigorous and reliable. 

\begin{table}[htb]
  \caption{Performance comparison with our proposed model by masking different parts of input and only using the remainder as input. Here "Summary", "SI" and "Video" indicates product core feature summary, structured information and product displaying video respectively.}
  \label{input_result}
  \centering
  
 \resizebox{\linewidth}{!}{
  \begin{tabular}{l|c|c|c|c|c|c}
    \toprule
    Input  & Bleu1 & Bleu2 &Bleu3 &Bleu4& Rouge\_L& CIDEr \\
    \midrule
    SI $\&$ Video& 22.8&14.8  & 6.9 & 5.5   & 22.2  & 25.3   \\
    Summary $\&$ Video& 19.5& 9.4 & 4.5 & 3.1  & 16.4 & 15.7  \\
    Video &15.9& 6.4 & 3.4  & 2.1  & 15 &  13.2       \\
    Summary $\&$ SI & 22.0&13.8  & 5.8 & 4.9   & 20.6  & 23.7   \\
    \bottomrule
  \end{tabular}}
\end{table}

2) The lack of a commodity core feature summary or displaying video will impair the foundation of generated text. In addition, the structured information is like a knowledge base, which can promote inference and judgment to generate appropriate copywriting.

\textbf{Conceptual Operation.} Considering that writing product descriptions in real life often involves a great number of various words, which makes it hard for the model to identify and remember its feature when facing a new word, such as new brand name. The predecessor's approach tend to use as much corpus and large model parameters as possible, which brings huge difficulties to natural language generation. In this case, we proposed the Conceptualization Operation. As shown in Table~\ref{CO_compare} , we conduct ablation experiments about Conceptualization on the Chinese and English datasets. As for models without conceptual operations, we use unconceptualized captions as the ground truth to train. We directly input unordered structured words for the input of the model. Experiments have proved that the Conceptualization operation can indeed bring a significant effect improvement, because this method can conceptualize and extract information from complex information in the dataset, and thus highly conceptualize network features. We expect this discovery to inspire the community.

\begin{figure*}[htb]
\setlength{\abovecaptionskip}{-0.3pt}
\centering
\scalebox{0.5}{
\includegraphics[width=2.0\textwidth]{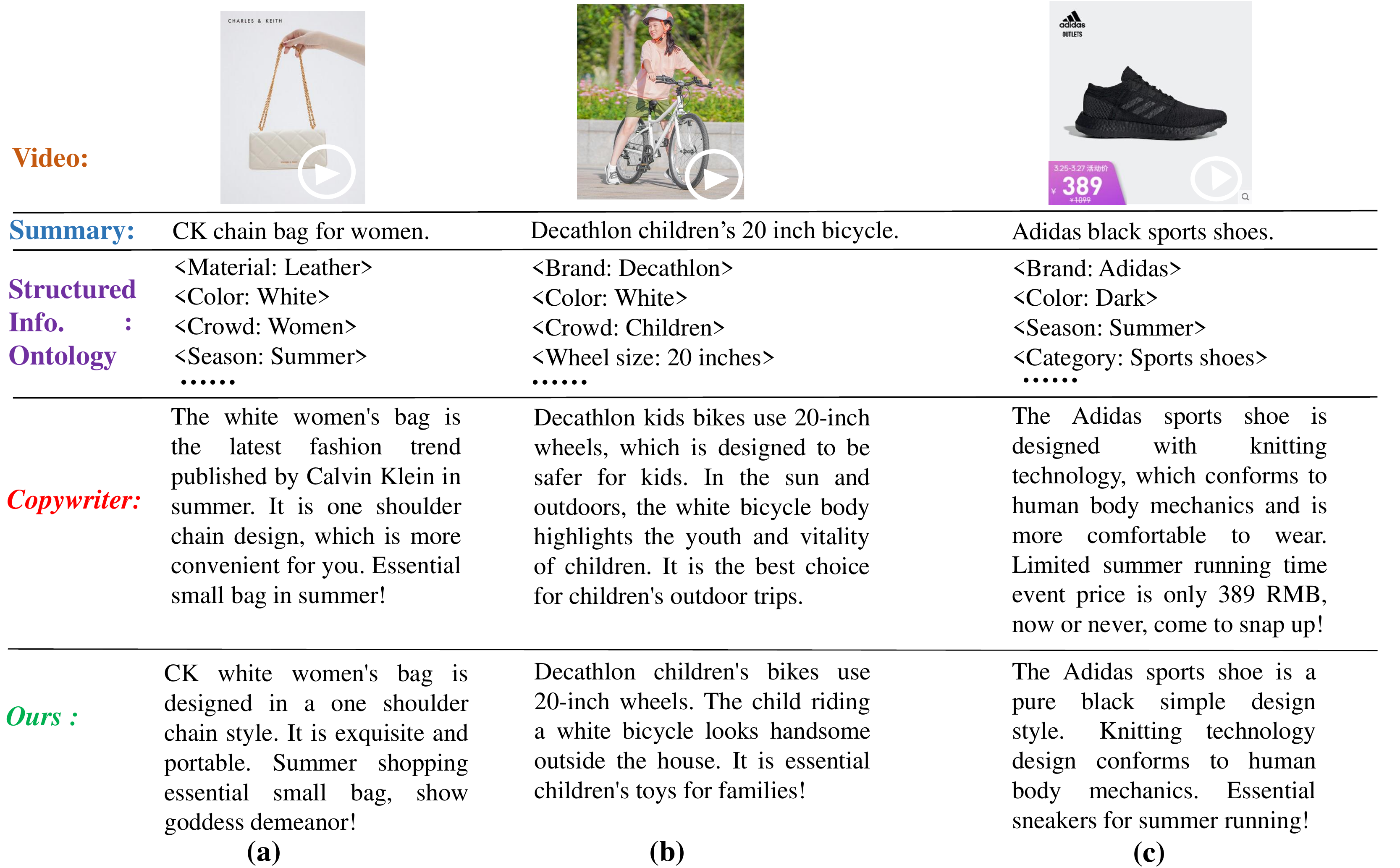}}
\caption{Some example results generated by our methods.  }
\label{fig:last}
\end{figure*}

\begin{table}[htb]
\setlength{\belowcaptionskip}{-0.2cm}
  \caption{Performance comparison of whether our proposed model has conceptual operations (CO).}
  \label{CO_compare}
  \centering
 \resizebox{\linewidth}{!}{
  \begin{tabular}{l|c| c| c| c| c| c}
    \toprule
    Operation  & Bleu1 & Bleu2 &Bleu3 &Bleu4& Rouge\_L& CIDEr \\
    \midrule
    ours w/o CO (en) &23.8  &15.4   &8.1    &6.4  & 24.2 & 27.3\\
    ours w/o CO (zh) &9.9  &5.5  &2.8  &1.5  &10.1  &12.4 \\
    ours w/ CO (en)& \textbf{25.0} & \textbf{16.6} & \textbf{9.6} &\textbf{7.2}& \textbf{25.3} &\textbf{29.1} \\
    ours w/ CO (zh)& \textbf{11.6} & \textbf{6.5} & \textbf{4.4} &\textbf{2.2}& \textbf{12.5} &\textbf{15.3} \\
    \bottomrule
  \end{tabular}}
\end{table}
\section{Evaluation Metric}
\subsection{Hard Homologous Metric}
\label{metric}
As shown in Table~\ref{compare_sota}, we firstly adopt three common NLG metrics. BLEU \cite{papineni2002bleu}, ROUGE\_L \cite{lin2004rouge}, CIDEr \cite{vedantam2015cider}. However, in reality application, as shown in Figure~\ref{fig:last},  we found that, the advertisement copywriting has the characteristics of flexible language style and rich vocabulary, and faithfulness\cite{2022Faithfulness} is particularly important. In terms of faithfulness, product advertising especially focuses on key information such as product brand and color, which is not fully reflected in the above indicators. For example, for Adidas shoes, the model can be lazy and output Nike brand because there is a similar training corpus in the training dataset, which is a common and serious error. To address this problem, we proposed a hard homologous metric.  In terms of accuracy, we traverse the attribute words in each ground truth (dataset has marked the key ontology of each value word), and compare with generation to calculate the proportion of correct words in generated words. In terms of error rate, according to figure~\ref{fig:statisc} statistical data and realistic requirements, we successively select brand, color, material, people, time and season as the five key labels as top-5 core attribute words. If they are inconsistent, they will be regarded as errors.  In the meantime, we'll call the rest unknown. Under such hard homologous metric, significant statistical analysis results were shown in Table~\ref{metric_table} for the faithfulness of product advertisements copywriting.
\begin{table}[htb]
\setlength{\abovecaptionskip}{-0.5pt}

\caption{The results of our propose hard homologous metric.}
\label{metric_table}
\centering
\resizebox{\linewidth}{!}{
\begin{tabular}{ll|l| l |l}
\toprule
                                                                                                     &                & Correct Rate    & Erro Rate       & Unknown         \\ \hline
\multicolumn{1}{l|}{\multirow{2}{*}{Ours}}                                                           & w/ conception  & \textbf{18.7\%} & \textbf{20.2\%} & 61.1\% \\
\multicolumn{1}{l|}{}                                                                                & w/o conception & 13.8\%          & 27\%            & 59.2\%          \\ \hline
\multicolumn{1}{l|}{\multirow{2}{*}{\begin{tabular}[c]{@{}l@{}}NLG\\ (CPM)\end{tabular}}}            & w/ conception  & 14.6\%          & 23.1\%          & 62.3\%          \\
\multicolumn{1}{l|}{}                                                                                & w/o conception & 9.8\%           & 30\%            & 60.2\%          \\ \hline
\multicolumn{1}{l|}{\multirow{2}{*}{\begin{tabular}[c]{@{}l@{}}Video caption\\ (NACF)\end{tabular}}} & w/ conception  & 10.9\%          & 29\%            & 60.1\%          \\
\multicolumn{1}{l|}{}                                                                                & w/o conception & 5\%             & 38\%            & 57\%            \\ \bottomrule
\end{tabular}}
\vspace{-0.6cm}
\end{table}

\subsection{Human Assessment}
It is well-known that the human evaluation metrics\cite{van2020human} for video captioning are required due to the inaccurate evaluation by automatic metrics. We especially focus on advertising generation, which depend on human aesthetics. So we invite the people involved in the data annotation and new advertising slogan designers to conduct the human evaluation. We select 200 samples randomly and each appraiser evaluates each of these 200 samples to reflect the performance of our model by rating whether the copywriting generated by our model can be used as a description of the product. As shown in Table~\ref{human_assess}, the advertisement copywriting generated by our model has a certain degree of pass rating, whose results can be approved by people. Therefore, this is also acceptable that our experiments on Table~\ref{reslut_compare} did not achieve high scores for mechanical evaluation indicators. We also test the CPM and human performance. The human performance test results were generated by the merchant copywriters.

\begin{table}[h]
\setlength{\abovecaptionskip}{-0.1pt}
  \caption{The pass rate comparison results  of the human evaluation, reflecting the proportion of the 200 reality application test examples where the model generated caption could be used as a product description that describes the reasonableness of the generated caption. Annotators are from the dataset annotation and persons are from the frequent online shopping masses.}
  \label{human_assess}
  \centering
 \resizebox{\linewidth}{!}{
  \begin{tabular}{c|c| c| c| c| c| c}
    \toprule
    \  & Annotator 1 & Annotator 2 &Annotator 3 &Person 1 & Person 2 &Person 3 \\
    \midrule
    Ours &\bf 42\%  &\bf44\%   &\bf43\% &\bf50\%  &\bf56\%   &\bf53\%\\
   \hline
    CPM &30\% &23\% &27\% & 40\% &47\% &39\%\\
    \hline
    Human &74\%  &77\% &79\% &90\% &81\% &89\%\\ 
    \bottomrule
  \end{tabular}}
\vspace{-0.6cm}
\end{table}

\section{Conclusion and Future Work}
\label{conclusion}

This work sets out to provide an e-commercial multimodal multi-structured advertisement copywriting dataset, E-MMAD, which is one of the largest video captioning datasets in this field. Based on E-MMAD, we also present a novel task: e-commercial multimodal multi-structured advertisement copywriting generation, and propose a baseline method on the strength of multi-structured information reasoning to solve the realistic demand. We hope the release of our E-MMAD would facilitate the development of multimodal generation problems. However, there still exist limitations about our dataset and method that should be acknowledged as shown in Figure~\ref{fig:last}. We cannot identify the price information of the video in \textbf{(c)}, which may require video OCR or ASR technology. Moving forward, we are planning to extend E-MMAD to better performance and more diversified tasks by exploring new model structures, fine-grained and so on.

\bibliographystyle{ACM-Reference-Format}
\bibliography{sample-base}

\clearpage
\appendix


\section*{Supplementary material}

# A  DATASET SUPPLEMENT

There are the data examples and standards as shown. We have built a special website for E-MMAD, which contains the dataset download, code link, and so on. We will open the experience function to the public.

Please refer to our video for a detailed introduction.

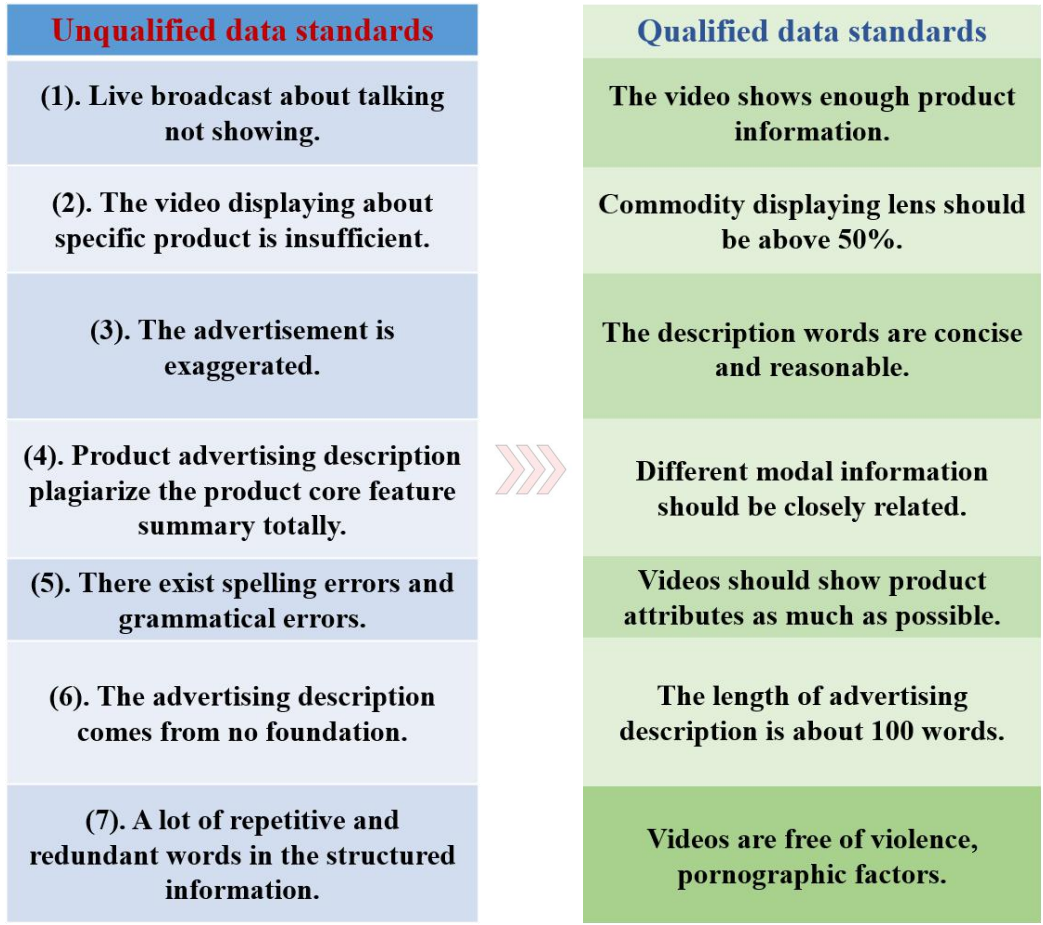

Figure 1: The manual filtering rules. The core rule of video screening is to show the visual and spatial features of products as much as possible. Our hope multimodal information is closely related to advertising copywriting. We also show some of the properties of qualified data.

*Unqualified Data Example*

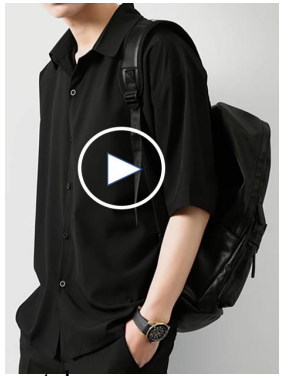 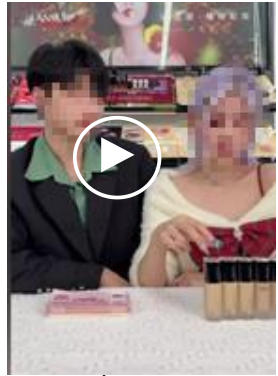 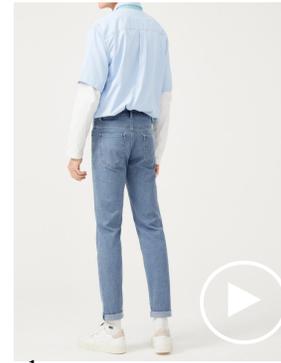

| | watch | cosmetic | shoe |
|---|---|---|---|
| *Reason:* | It's hard to focus on the feature of the hand watch. It violates rule (2). | The cosmetics in the lower right corner are not obvious visually. It violates rule (1), (2). | Jeans? Blue shirt? It's hard to focus the white shoe. It violates rule (2). |

**Figure 2: There are some unqualified data examples. The visual feature plays an importance role in multimodal task.**

*Qualified Data Example*

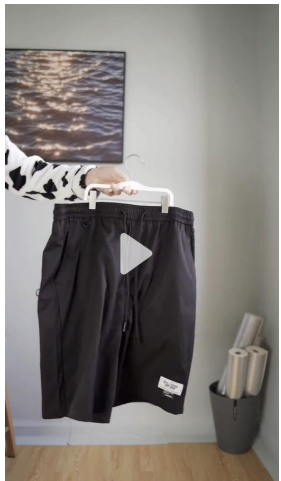 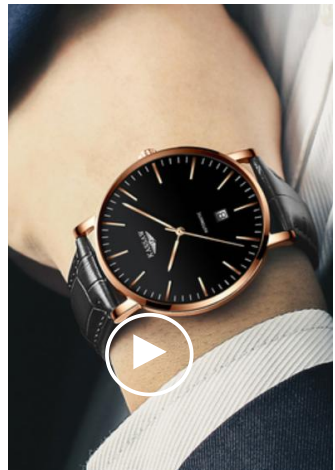 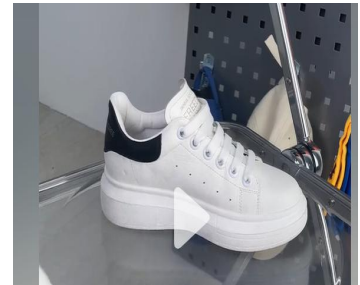

(1)                         (2)                         (3)

*Ad :*    (1) Semir sports shorts designed in pure black, are simple style for men. This is a comfortable loose version, allowing you to enjoy the comfort of sports without restraint. Wear it for a cool summer!
(2) This is a beautiful and atmospheric mechanical watch. Its strap is designed with stainless steel material, which makes the watch more textured. And the watch also adopts a simple dial design. The light on the wrist, the watch achieves masculine charm.
(3) This is Anta's new 2022 small sneakers. The upper is made of high-quality leather fabric, which is breathable and comfortable, soft and wear-resistant, and durable. With a solid color system, it is low-key and restrained, beautiful and elegant. With fine workmanship, it shows the quality of shoes. Coupled with the preferred lining, it is comfortable and comfortable, bringing a more excellent wearing experience to the feet.

**Figure 3: There are some qualified data examples. We hope to see the wonderful advertisement copywriting, which is closely related to multimodal information.**

 

\end{document}